\documentclass[letterpaper]{article} 
\usepackage{aaai25}  
\usepackage{times}  
\usepackage{helvet}  
\usepackage{courier}  
\usepackage[hyphens]{url}  
\usepackage{graphicx} 
\usepackage{natbib}  
\usepackage{caption} 
\usepackage{algorithm}
\usepackage{algorithmic}
\usepackage{booktabs}
\usepackage{subcaption}
\usepackage{amsmath} 
\usepackage{lipsum}  

\usepackage{newfloat}
\usepackage{listings}
\DeclareCaptionStyle{ruled}{labelfont=normalfont,labelsep=colon,strut=off} 
\floatstyle{ruled}
\newfloat{listing}{tb}{lst}{}
\floatname{listing}{Listing}
\title{Large-scale School Mapping using Weakly Supervised Deep Learning for Universal School Connectivity}
\author{
    Isabelle Tingzon,
    Utku Can Ozturk,
    Ivan Dotu}
\affiliations{
    United Nations Children's Fund (UNICEF)\\
    itingzon@unicef.org, uozturk@unicef.org, jdoturodriguez@unicef.org}

\usepackage{bibentry}

\begin{document}

\maketitle

\begin{abstract}

Improving global school connectivity is critical for ensuring inclusive and equitable quality education. To reliably estimate the cost of connecting schools, governments and connectivity providers require complete and accurate school location data – a resource that is often scarce in many low- and middle-income countries. To address this challenge, we propose a cost-effective, scalable approach to locating schools in high-resolution satellite images using weakly supervised deep learning techniques. Our best models, which combine vision transformers and convolutional neural networks, achieve AUPRC values above 0.96 across 10 pilot African countries. Leveraging explainable AI techniques, our approach can approximate the precise geographical coordinates of the school locations using only low-cost, classification-level annotations. To demonstrate the scalability of our method, we generate nationwide maps of school location predictions in African countries and present a detailed analysis of our results, using Senegal as our case study. Finally, we demonstrate the immediate usability of our work by introducing an interactive web mapping tool to streamline human-in-the-loop model validation efforts by government partners. This work successfully showcases the real-world utility of deep learning and satellite images for planning regional infrastructure and accelerating universal school connectivity.
\end{abstract}

\section{Introduction}


Globally, approximately 2.2 billion children and young people -- two-thirds of the world's youth -- do not have access to the internet \cite{unicef2020many}. The absence of internet connectivity not only limits children's opportunities for online education but also prevents them from developing the digital skills needed to compete in the modern economy. Disparities in school connectivity can exacerbate existing inequities and widen the gap in educational outcomes for children with and without internet access. To help bridge the digital divide, the United Nations Children’s Fund (UNICEF) and International Telecommunication Union (ITU) launched Giga, a global initiative to connect every school to the internet by 2030. To reach this target, governments and connectivity providers require complete and accurate school location data to reliably estimate the cost of connecting schools and strategically allocate their financial resources. 

However, while governments generally have comprehensive records of the schools in their national register, these records often lack geographical coordinates, especially in developing countries. For example, government partners in Senegal estimate that approximately 20\% of school geolocations are missing from their official dataset. Meanwhile, only about 7,000 out of the estimated 33,000 schools in Kenya have corresponding GPS coordinates \cite{giga2024,kenya}. These unmapped schools are often located in rural and remote areas, meaning that without accurate data, governments and internet service providers risk overlooking the most vulnerable child populations.



To address these challenges, we look towards deep learning and satellite imagery to close critical gaps in school location data. Previous studies have shown that despite variations in school structures across countries, schools typically have identifiable overhead signatures that make them distinguishable from high-resolution satellite imagery \cite{maduako2022automated,yi2019towards}. However, extracting highly local school location information typically requires costly bounding box or pixel-level annotations, which can be challenging to acquire on a global scale \cite{fu2021detection,fu2022feature}. 

This study improves upon previous works by introducing a weakly supervised deep learning approach that approximates the precise geographical coordinates of school locations using only low-cost, classification-level annotations \cite{lee2021scalable}. We began by developing a pipeline to create country-level school mapping datasets by integrating information from various public data sources, including OpenStreetMap (OSM), Overture Maps, and GigaMaps. Leveraging model ensembling techniques that combine transformer-based models and convolutional neural networks (CNNs), we trained school classification models using satellite images, achieving AUPRC scores above 0.96 across 10 pilot African countries. We then used explainable AI (XAI) to further localize the schools within the images. We also examined how model performance varies between urban and rural subregions and compared regional models (trained on data from multiple countries) with local models (trained on data from a single country).

Using our best-performing models, we demonstrate the viability and scalability of our approach by generating nationwide maps of school locations for selected countries in Africa and present a detailed analysis of the results. Finally, we introduce an interactive web mapping tool to streamline human-in-the-loop model validation efforts, collaborating closely with government partners to accelerate the discovery of previously unmapped school locations.


\section{Related Works}
Previous works have demonstrated the potential of deep learning for automated school mapping. Most relevant to our work are the studies by \citet{maduako2022automated} and \citet{yi2019towards}, which explore the use of CNNs for tile-based classification of schools from high-resolution satellite images. Our work improves upon these previous studies in several ways. For one, prior works typically sampled non-school tiles from unpopulated areas such as forests, deserts, and bodies of water \cite{maduako2022automated, yi2019towards}. However, this can lead to artificially inflated model performance as these uninhabited non-school tiles, which are easily distinguishable from school tiles, do not represent the complex, built-up environments where the model will be deployed in real-world settings. In contrast, we focus the deployment of our model in built-up areas and ensure that non-school tiles for training are also sampled from within these regions. 

Moreover, in addition to the traditional CNNs commonly used in past literature, we employ vision transformers (ViTs) which have emerged as a promising technique for satellite image classification \cite{dosovitskiy2021image,bazi2021vision}. Self-attention mechanisms enable ViTs to capture long-range dependencies between patches within an image. This makes ViTs potentially well-suited for learning the relationships between different components of a school -- including playgrounds, track and field ovals, basketball courts, open fields, and grouped building structures -- as seen from overhead imagery \cite{maduako2022automated}. 


More recently, studies by \citet{fu2021detection,fu2022feature} have explored the use of object detection models to extract more granular school location information in China. Similarly, benchmark datasets like the Functional Map of the World (fMoW) and the Urban Building Classification (UBC) dataset have included educational facilities and school buildings as categories among several fine-grained building types for building detection and classification \cite{christie2018functional, huang2022urban}. However, studies employing object detection or instance segmentation models typically require costly bounding box or pixel-level annotations, which can be time-consuming and labor-intensive to acquire. 

Here, we present a weakly supervised deep learning approach that requires only classification-level annotations for granular school localization. This method builds upon previous works that have successfully leveraged XAI techniques to approximate the precise lat-lon coordinates of objects from satellite images, including brick kilns and industrial poultry operations  \cite{lee2021scalable,handan2019deep}. 

\section{Data}

\subsection{Data preprocessing}
 We began with official school data from government partners for 10 African countries: Benin (\textbf{BEN}), Botswana (\textbf{BWA}), Ghana (\textbf{GHA}), Kenya (\textbf{KEN}), Malawi (\textbf{MWI}), Namibia (\textbf{NAM}), Rwanda (\textbf{RWA}), Senegal (\textbf{SEN}), South Sudan (\textbf{SSD}), and Zimbabwe (\textbf{ZWE}). Each government-acquired dataset contains the school names and GPS coordinates, made accessible via GigaMaps \cite{giga2024}. We augmented each country-level dataset with school point-of-interest (POI) information from Overture Maps and OSM, retrieved using DuckDB and the Overpass API, respectively. 
 
As this study focuses primarily on primary and secondary schools, we excluded schools containing keywords related to early childhood education (e.g. preschool, kindergarten), tertiary education (e.g. university, college), sports academies (e.g. swimming, taekwondo), and other types of educational institutions. Next, we identified groups of duplicate points (i.e. coordinates that represent the same school location) by creating 150 m buffers around each point and aggregating those with overlapping buffers. From each group, we retained a single point and discarded the rest to ensure that the minimum distance between any two remaining points is 300 m. To remove erroneous points in unpopulated areas (e.g. forests, deserts, grassland), we used the Global Human Settlements Layer (GHSL) \cite{pesaresi2023ghs} along with Microsoft Building Footprints \cite{microsoft}, and Google Open Buildings \cite{sirko2021continental}, rasterized to 10 m resolution GeoTIFFs. For each school location in the dataset, we calculated the number of settlement pixels within a 150-meter buffer around the school. Points with zero human settlement pixels within their buffer areas across all three settlement datasets were subsequently discarded. 

\subsubsection{Non-school samples} To generate our set of negative samples, we queried the locations of non-school POIs such as hospitals, churches, hotels, and office buildings from OSM and Overture Maps. Due to the scarcity of POI data in low- and middle-income countries, the number of positive school samples can  exceed the number of negative samples from OSM and Overture. However, in real-world settings, we expect the number of non-school image tiles to be much larger than the number of school image tiles. Without a sufficient set of negative samples, the model may fail to capture the full variability of non-school areas, leading to poor generalizability. We therefore increased the number of non-school data points by randomly sampling points from populated areas, with the assumption that a vast majority of these points are non-school locations. For consistency, we adopted a fixed imbalanced ratio of 1:2 (positive to negative) across the 10 countries to achieve a higher degree of sample variability among non-school tiles. To prevent data leakage, we ensured that the sampled points were spaced a minimum distance of 300 meters apart. 

Finally, we applied the same data cleaning pipeline for non-school locations, with the additional step of removing non-schools that were within 300 meters of known school locations. This was done to ensure that no known school building appears in the periphery of non-school satellite images. For each school and non-school location in our country-level datasets, we downloaded 300 x 300 m, 500 x 500 px high-resolution satellite images from Maxar with a spatial resolution of 60 cm/px, centered on the corresponding GPS coordinate \cite{maxar2024}. 

\subsubsection{Sources of noise} Combining data across multiple sources allows us to create a rich and diverse dataset. However, public records and crowd-sourced information can introduce noise and undermine the correctness of the school mapping dataset. We identify potential sources of noise for each class in our dataset as follows: 
\begin{itemize}
    \item \textbf{School location noise.} The accuracy of the school's GPS coordinates can vary depending on the mode of data collection. In many cases, the coordinates are not recorded at the school building itself but in the surrounding area, such as at the outer gates, along the nearest road, or in nearby open fields. Moreover, due to the absence of precise locational data, governments would sometimes position the school's coordinates at the center of its administrative boundaries instead of at the actual school location. From visual inspection, we observed a considerable number of school coordinates that were located several hundred meters away from the actual school building, which poses a challenge to the accuracy of the labels.
    \item \textbf{Non-school location noise.} Aerial images of negative samples may contain school buildings that are not among the known schools in our school mapping dataset. While the goal of this work is to ultimately discover these unmapped school locations, these schools may inadvertently be included as noise among non-school samples. 
\end{itemize}

To improve the correctness of the data, we manually reviewed each satellite image of known school locations and removed images where the school appeared to be either absent from the image or indistinguishable from surrounding buildings. We also resolved location-related discrepancies by manually repositioning the GPS coordinates of schools located more than 300 m away from the actual school building, based on auxiliary information from the Google Satellite Hybrid base map. \\

\subsection{Data split} 
Understanding how model performance varies across subpopulations is important in mitigating bias and ensuring fairness \cite{mitchell2019model}. In this study, we consider the degree of urbanization to be a relevant factor for evaluating our school classification model and report the disaggregated model performance with respect to urban and rural subgroups. We assigned urban/rural labels to each data point using the GHSL-SMOD L2 product \cite{schiavina2022ghs}. GHSL-SMOD classifies 1 km$^2$ grid cells into clusters based on the degree of urbanization. We consider 2 main clusters: (1) urban domain, which comprises urban center grid cells, dense urban grid cells, semi-dense urban grid cells, and suburban or per-urban grid cells; and (2) rural domain, which consists of rural cluster grid cells, low-density grid cells, and very low-density grid cells. 

Each country-level dataset is split into a training set (80\%), validation set (10\%), and test set (10\%). The data is split using stratified random sampling of non-overlapping 300 x 300 m tiles such that we preserve the ratio of positive and negative samples from the overall set of images and the ratio of urban and rural samples per class. This strategy ensures representation of both urban and rural areas in each split. Note that for country-level experiments, the problem is formulated as a spatial interpolation task \cite{wadoux2021spatial,rolf2023evaluation}. We present in Table \ref{tab:class-dist} the per-country class distribution and the urban/rural distribution per class.


\subsubsection{Generalizability experiments} Next, we evaluate the cross-country generalization of each country-specific model, i.e. we seek to determine how well a model trained in country $A$ would perform in country $B$. Following \citeauthor{beery2022auto}, we perform all possible cross-training combinations, training on each country's training set and evaluating on every other country's designated test set. Lastly, to assess whether transnational, geographically diverse training data improves model performance, we train a regional model for Africa using the combined training datasets from all 10 countries, hereby referred to as the \textit{regional} dataset, and test on the individual test sets of each country. 
\begin{table}[t]
\centering
\resizebox{\linewidth}{!}{%
    \begin{tabular}{lrrrcrrrcr}
    \toprule
    & \multicolumn{3}{c}{\textbf{School}} && \multicolumn{3}{c}{\textbf{Non-school}} && \textbf{Total}  \\
    & \textbf{Urban} & \textbf{Rural} & \textbf{Total} && \textbf{Urban} & \textbf{Rural} & \textbf{Total} && \textbf{} 
    \\
    \cline{2-4} \cline{6-8} \cline{10-10}
    \noalign{\vspace{1ex}}
    \textbf{BEN} &  1,930 & 2,104 & 4,034 && 4,060 & 4,008 & 8,068 && 12,102 \\
    \textbf{BWA}   
    & 411 & 457 & 868 && 933 & 803 & 1,736 && 2,604
    \\
    \textbf{GHA}  
    &  3,013 & 4,412 & 7,425 && 8,357 & 6,493 & 14,850 && 22,275
    \\
    \textbf{KEN}
    & 3,642 & 1,432 & 5,074 && 7,265 & 2,883 & 10,148 &&  15,222
    \\
    \textbf{MWI}
    & 2,077 & 1,034 & 3,111 && 3,996 & 2,226 & 6,222 && 9,333
    \\
    \textbf{NAM}
    & 318 & 1,374 & 1,692 && 701 & 2,683 & 3,384 && 5,076
    \\
    \textbf{RWA}
    & 2,473 & 78 & 2551 && 4,737 & 365 & 5,102 && 7,653
    \\
    \textbf{SEN}
    & 1,955 & 4,443 & 6,398 && 5,655 &  7,141 & 12,796 && 19,194
    \\
    \textbf{SSD}
    & 562 & 1,018 & 1,580 && 802 & 2,358 & 3,160 && 4,740
    \\
    \textbf{ZWE}
    & 698 & 3,233 & 3,931 && 1,675 & 6,187 & 7,862 && 11,793
    \\
    \bottomrule
    \end{tabular}}
    \caption{Class distribution and urban/rural distribution across the 10 African countries.}
    \label{tab:class-dist}
\end{table}

\section{Methods}
Our pipeline comprises a two-stage process that involves (1) training an image classifier to determine whether or not a 300 x 300 m satellite image contains a school and (2) using class activation maps (CAMs) to further localize the geographic coordinates of the school within the image. 

\subsection{Image-level prediction} 
 For the image classification task, we experimented with different model architectures, including three variants of ViT (base, large, and huge) \cite{dosovitskiy2021image}, three variants of Swin Transformer V2 (SwinV2) (tiny, small, base) \cite{liu2022swin}, and three variants of ConvNext (small, base, and large) \cite{liu2022convnet}. Per country, we also implemented an ensemble approach, termed VSC-Ensemble, that calculates the mean of the softmax vectors of the best-performing variant of each type of model architecture (ViT, SwinV2, and ConvNext) \cite{sivasubramanian2024transformer}. 
 
 For model training, we adopted a transfer learning approach wherein all models were initially pre-trained on the ImageNet dataset \cite{deng2009imagenet} and then fine-tuned on the designated training sets per country using a cross-entropy loss with label smoothing set to 0.1 for regularization. We resized the images to 224 x 224 px and implemented data augmentation for the training set in the form of vertical and horizontal flips as well as random rotations. For ViT and SwinV2 models, we used an initial learning rate (LR) of $1e^{-5}$, and for ConvNext, we approximated the optimal initial LR using the LR range test by varying the LR between $1e^{-3}$ and $1e^{-6}$ over 1,000 iterations \cite{silva2024lrfinder,smith2017cyclical}. The LRs were set to decay by a factor of 0.1 after every 7 epochs of no improvement. Across all models, we used an Adam optimizer, a batch size of 32, and a maximum number of epochs of 60, with early stopping if the LR fell below $1e^{-7}$.  All models were trained in a single iteration in high-performance computing (HPC) environments using Python 3.10.13 on a Linux 4.18.0 operating system with NVIDIA A40 and NVIDIA A100 80GB PCIe GPUs. 

\subsection{School localization}
We employed a weakly supervised approach wherein the model learns from the image classification task to perform the more difficult task of school localization, i.e. finding where within the image the school is located. We leverage XAI techniques, specifically CAMs, to assign importance scores to the pixels in the satellite image that contribute most to the school prediction. From the CAMs, we can then convert the xy coordinates of the most important pixel to a lat-lon coordinate to approximate the geographic coordinates of the school. The approach is considered weakly supervised as it does not require exact annotations of the school location (e.g. bounding boxes, pixel masks), which can be labor-intensive and time-consuming to generate \cite{lee2021scalable}. 

\begin{figure}
    \centering
    \includegraphics[width=1\linewidth]{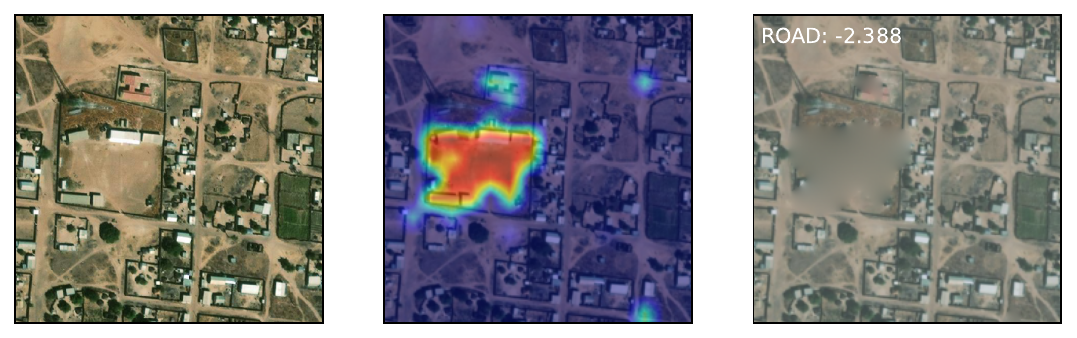}
    \caption{(Left) 300 x 300 m (500 x 500 px) satellite image tile of a school in Senegal. (Middle) GradCAM outputs. (Right) Perturbation of the top 10\%  of pixels using ROAD.}
    \label{fig:cam-results}
\end{figure}

We experimented with different pixel attribution methods, including GradCAM \cite{selvaraju2017gradcam}, GradCAMElementWise \cite{pillai2021explainable}, 
GradCAM++ \cite{chattopadhay2018gradcampp}, HiResCAM \cite{draelos2021hirescam}, EigenCAM \cite{muhammad2020eigencam}, and LayerCAM \cite{jiang2021layercam} as implemented in the Pytorch library for CAM methods \cite{jacobgilpytorchcam}. Across all models, we chose the normalization layer before the final block as our target layer. Figure \ref{fig:cam-results} depicts an example of GradCAM outputs for a satellite image in Senegal. 

\section{Results and Discussion}
\subsection{Image-level school classification} Because the optimal decision threshold (i.e. the probability level at which to classify an image as containing a school) can vary based on the end user's error tolerance, we assessed the model performance at all possible decision thresholds between 0 and 1. Specifically, we measured precision and recall at every threshold $\tau$, wherein a school prediction is considered correct if a satellite image of a school receives a probability score $> \tau$. We report our primary performance metric as the area under the precision-recall curve (AUPRC) and select the model that achieves the highest AUPRC on the test set for each country. A comparison of the different variants of ViT, SwinV2, and ConvNext models across the 10 different countries is presented in Table \ref{tab:model-results}. 

Our results indicate that for  image classification, transformer-based models (i.e. ViT, SwinV2) generally outperform CNNs for a majority of the countries. Unlike CNNs that learn feature representation in a hierarchical manner (i.e. earlier layers detect simple patterns while latter layers detect more complex features), transformers can model global dependencies in a single shot using self-attention mechanisms. Since schools in Africa typically consist of multiple components (e.g. buildings, open fields, playgrounds), it is possible that vision transformers, with their ability to capture long-range dependencies, are better at retaining critical contextual information that may be lost in the latter layers of CNNs. 

\begin{figure}
    \centering
    \includegraphics[width=\linewidth]{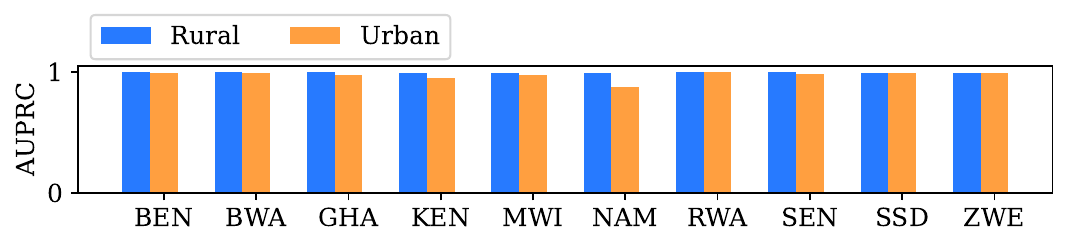}
    \caption{Per-country AUPRC by urban/rural subgroup.}
    \label{fig:rurban-auprc}
\end{figure}

However, ultimately, the best-performing models across all 10 countries are a combination of ViT, SwinV2, and ConvNext models, i.e. VSC-Ensemble, demonstrating how the simple technique of model ensembling can significantly improve model performance. We also present in Figure \ref{fig:rurban-auprc} the AUPRC of the best-performing local models, disaggregated by urban/rural subgroups. We generally find that the models perform equally well or slightly worse in urban areas compared to rural areas. This is likely due to increased opportunities for errors in these environments, e.g. school-like structures such as hospitals, mosques, and government buildings are more prevalent in city centers, which can increase the likelihood of false positives. 

\begin{table*}[t]
    \centering
    \begin{tabular}{lccccccccccc}
    \toprule
    \textbf{Model} & \textbf{BEN} & \textbf{BWA} & \textbf{GHA} & \textbf{KEN} & \textbf{MWI} & \textbf{NAM} & \textbf{RWA} & \textbf{SEN} & \textbf{SSD} & \textbf{ZWE} & \textbf{Regional}\\
    \midrule
    ViT-Base  & 0.971 & 0.984  & 0.920 & \underline{0.906} & 0.958 & 0.952 & 0.960 & 0.961 & 0.925 & 0.960 & 0.952\\
    ViT-Large  & 0.964& \underline{0.989} & 0.915 & 0.905 & 0.953 & 0.946 & 0.961 & 0.964 & 0.931 & 0.958 & 0.954\\
    ViT-Huge  & \underline{0.978} & 0.979 & \underline{0.930} & 0.906 & \underline{0.967} & \underline{0.955} & \underline{0.983} & \underline{0.980} & \underline{0.971} & \underline{0.971} & \underline{0.960} \\
    \midrule
    SwinV2-Tiny  & 0.965 & \underline{0.984} & 0.921 & 0.899 & 0.959 & 0.955 & \underline{0.982} & \underline{0.967} & \underline{0.964} & 0.955 & 0.958\\
    SwinV2-Small  & \underline{0.984} & 0.981 & \underline{0.931} & 0.904 & \underline{0.962} & \underline{0.949} & 0.978 & 0.953 & 0.932 & 0.958 & 0.960\\
    SwinV2-Base & 0.973 & 0.975 & 0.926 & \underline{0.910} & 0.961 & 0.952 & 0.966 & 0.960 & 0.957 & \underline{0.961} & \underline{0.963}\\
    \midrule
    ConvNext-Small & 0.953& 0.983 & \underline{0.929} & \underline{0.916} & 0.957 & 0.953 & 0.977 & 0.973 & 0.955 & 0.938 & 0.961\\
    ConvNext-Base & \underline{0.977} & 0.982 & 0.927 & 0.904 & \underline{0.969} & 0.951 & \underline{0.978} & 0.964 & \underline{0.964} & \underline{0.967} & \underline{0.961}\\
    ConvNext-Large & 0.945 & \underline{0.985} & 0.926 & 0.905 & 0.954 & \underline{0.954} & 0.977 & \underline{0.978} & 0.962 & 0.960 & 0.960\\
    \midrule
    VSC-Ensemble & \textbf{0.998} & \textbf{0.997} & \textbf{0.991} & \textbf{0.966} & \textbf{0.983} & \textbf{0.980} & \textbf{0.998} & \textbf{0.993} & \textbf{0.995} & \textbf{0.996} & \textbf{0.984}\\
    \bottomrule
    \end{tabular}
    \caption{A comparison of model performances (AUPRC). Each local model is trained on the designated training set and tested on the corresponding test set. The regional model is trained on the combined training sets of all countries and tested on the combined test sets. The best-performing variants per model architecture (ViT, SwinV2, and ConvNext) are underlined and ensembled via soft voting (VSC-Ensemble), and the best model performances overall are highlighted in bold.}
    \label{tab:model-results}
\end{table*}

\subsubsection{Cross-country generalization} We show in Figure \ref{fig:heatmap} a heatmap illustrating the cross-country generalizability of the best models across the 10 countries. We generally see a pattern where countries generalize well to neighboring countries (e.g. Benin and Ghana; Botswana, Namibia, and Zimbabwe). We also find that countries with a relatively small number of school samples (e.g. Botswana) do not generalize to other countries as well as those with a larger number of school samples (e.g. Ghana). This indicates that in addition to geographic proximity, quantity and representation (or lack thereof) are major factors affecting generalizability. 

\subsubsection{Regional models vs. local models} 
\begin{figure}
    \centering
    \includegraphics[width=\linewidth]{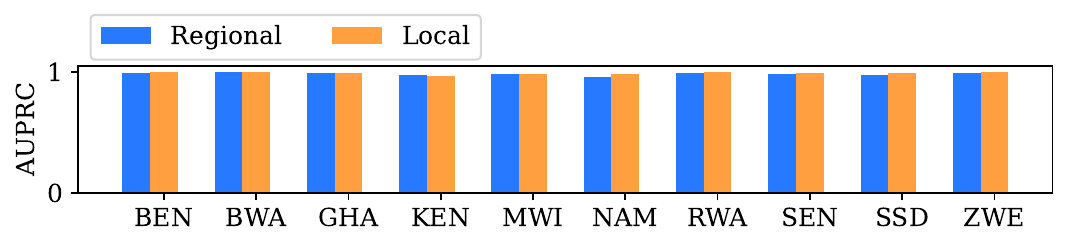}
    \caption{A comparison of the AUPRC of the best regional model versus the best local model per country.}
    \label{fig:regional-auprc}
\end{figure}

\begin{figure}
    \centering
    \includegraphics[width=\linewidth]{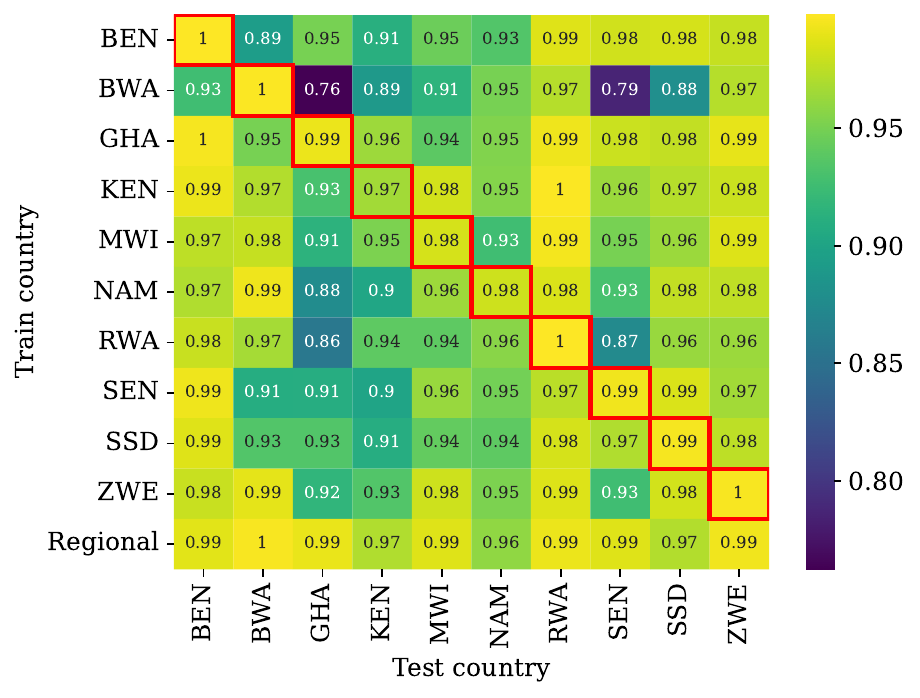}
    \caption{Pairwise train/test AUPRC.}
    \label{fig:heatmap}
\end{figure}

As shown in Figure \ref{fig:heatmap}, only the regional model generalizes well across all countries. However, when compared to the best country-specific model, the best regional model achieves similar performance scores as the best models trained on local data only, as shown in Figure \ref{fig:regional-auprc}. Consistent with past works \cite{maduako2022automated}, our results show that regional models trained on larger transnational, geographically diverse data do not provide significant performance improvements over models trained on local, country-specific data. These findings indicate that highly contextualized local data alone can already achieve strong model performance.


\subsection{Evaluating CAM methods for school localization}
 To assess the quality of the different pixel attribution methods, we measured the average confidence drop across all test set images after perturbing the pixels at the 90th percentile \cite{rong22consistent}. For the removal order, we used the Most Relevant First (MoRF) imputation strategy, which perturbs pixels in decreasing order of attention values \cite{jacobgilpytorchcam}. We employed the Remove and Debias (ROAD) evaluation framework, which uses noisy linear imputation to mitigate class information leakage arising from pixel masking \cite{rong22consistent}. Specifically, the ROAD framework solves a system of linear equations to replace each removed pixel with the weighted mean of its neighbors, as shown in Figure \ref{fig:cam-results}. Note that when the CAM method fails to produce meaningful results, the ROAD framework may perturb the entire image, causing a large drop in the confidence score. To address this issue, we used Canny edge detection to identify such instances and set their confidence drop values to zero \cite{canny1986computational}. 

For each country, we used the best-performing model (excluding the ensemble model) to compute the average confidence drop over the test set images following pixel perturbation using ROAD. The results shown in Table \ref{tab:cam-results} indicate that GradCAMs perform best for countries where the best-performing image classification model is ViT-Huge. For all other countries, GradCAM++ and GradCAMElementwise achieve the highest average confidence drop.

\begin{table*}[t]
    \centering
    \begin{tabular}{lccccccccccc}
    \toprule
    \textbf{CAM Method} & \textbf{BEN} & \textbf{BWA} & \textbf{GHA} & \textbf{KEN} & \textbf{MWI} & \textbf{NAM} & \textbf{RWA} & \textbf{SEN} & \textbf{SSD} & \textbf{ZWE}\\
    \midrule
    EigenCAM & 0.488 & 0.245 & 0.206 & 1.075 & 1.127  & 0.077 & 1.272 & 0.488 & 0.525 & 1.803\\
    EigenGradCAM & 1.027 & 0.280 & 0.617 & 1.162  & 1.329  & 0.592 & 1.826 & 1.584 & 0.705 & 1.686\\
    GradCAM & 0.532 & 0.643 & 0.518 & 0.954 & 0.210  & 0.568  & \textbf{2.057} & \textbf{2.061} & \textbf{0.792} & 2.025\\
    GradCAM++ & 0.175 & 0.268 & 0.111 & 0.565 & 0.581  & 0.127 & 1.965 & 1.776 & 0.697 & \textbf{2.083}\\
    GradCAMElementWise & \textbf{1.263} & \textbf{1.083} & \textbf{0.712} & \textbf{1.186} & \textbf{1.354}  & \textbf{0.950} & 1.724 & 1.678 & 0.698 & 1.826\\
    HiResCAM & 0.161 & 0.641 & 0.243 & 1.084 & 1.128 & 0.147 & 1.919  & 1.958 & 0.742 & 1.942\\
    LayerCAM & 0.177 & 0.534 & 0.190 & 0.793 & 0.708 & 0.164 & 1.922 & 1.965 & 0.743 & 1.956\\
    \bottomrule
    \end{tabular}
    \caption{A comparison of different pixel attribution methods based on the average confidence drop of the best local model over all the corresponding test set images following perturbation of the top 10\% of pixels using the ROAD framework.}
    \label{tab:cam-results}
\end{table*}

\subsection{Country-wide school predictions}
To generate nationwide maps of model-predicted school locations, we started by creating sliding windows of size 300 x 300 m, or 500 x 500 px within each country's boundary. Because schools can be split between images, we used an overlapping stride of 50\% to ensure that the schools are centered in at least one image.  To reduce the number of satellite images to download, we filtered out tiles that do not contain any human settlements based on the rasterized Microsoft Building Footprints dataset and the Google Open Buildings dataset \cite{sirko2021continental}. As an example, we downloaded approximately 500K satellite images for Senegal and 1.2M satellite images for Ghana. 

Per country, each satellite image tile is fed as input to the best-performing image classification model, and the best pixel attribution method is then used to approximate the school location coordinates. For simplicity, we only generated CAMs for images with probability scores above the threshold value $\tau^*$ that optimizes the F2 score of each country's validation set. Following \citet{robinson2022mapping}, we chose the F2 score, which assigns a higher weight to recall, as filtering out false positives is more feasible than adding back false negatives (i.e., missed schools) post-deployment. Table \ref{tab:fscore-results}  presents the optimal threshold used for CAM generation and the corresponding F2 score, precision, and recall of the test set per country.

\begin{figure}
    \centering
    \includegraphics[width=0.7\linewidth]{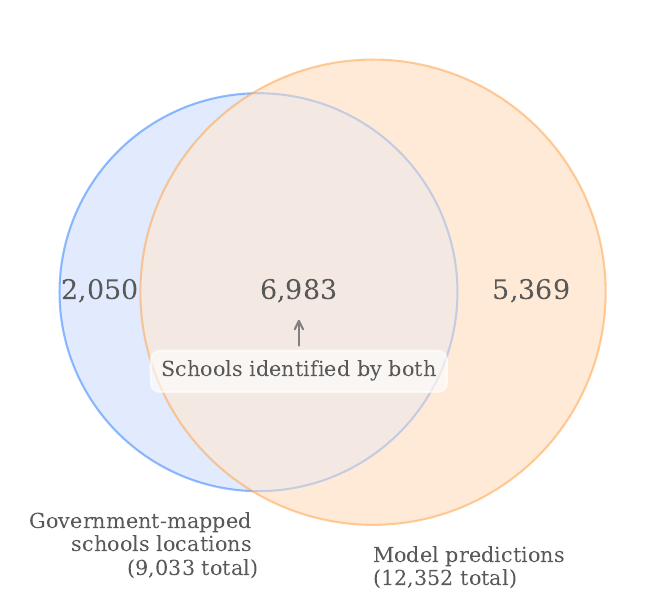}
    \caption{A comparison of model predictions and government data for Senegal ($\tau=0.5$ and $d=250 m$).}
    \label{fig:venn-diagram}
\end{figure}

Because it is possible for the same school to be detected in more than one image, we created a 50 m buffer around each predicted school coordinate and aggregated points with overlapping buffer areas. For countries with larger schools (e.g. Botswana) we increased the buffer size to 150 m. From each group of points, we retained the coordinate with the highest probability score. 

\begin{figure*}[t!]
    \centering
    \begin{subfigure}[t]{0.25\textwidth}
        \centering
        \includegraphics[width=\linewidth]{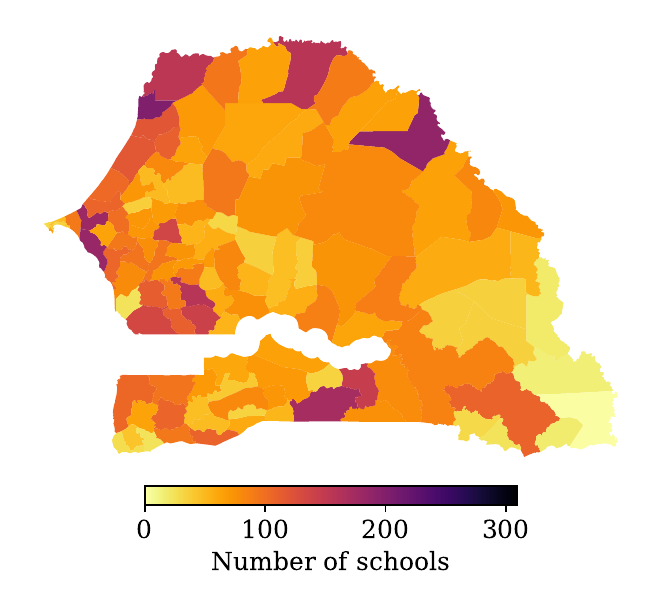}
        \caption{Government-identified schools}
    \end{subfigure}\hspace{0.05\textwidth}
    \begin{subfigure}[t]{0.25\textwidth}
        \centering
        \includegraphics[width=\linewidth]{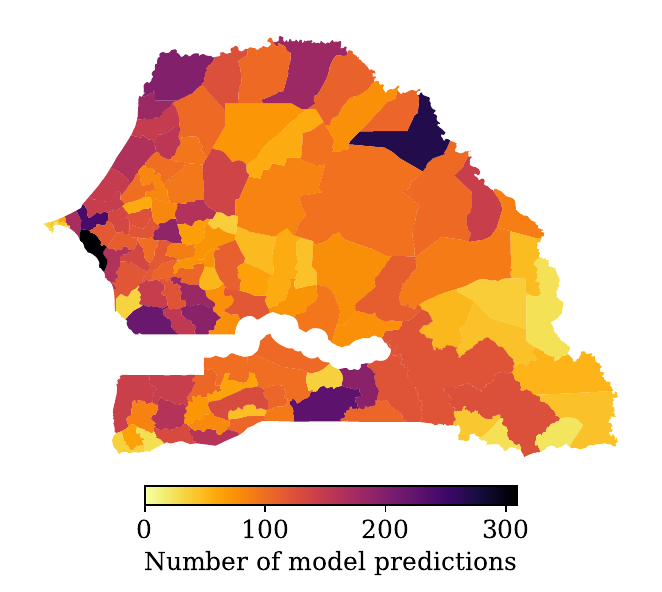}
        \caption{School predictions}
    \end{subfigure}\hspace{0.05\textwidth}
    ~ 
    \begin{subfigure}[t]{0.25\textwidth}
        \centering
        \includegraphics[width=\linewidth]{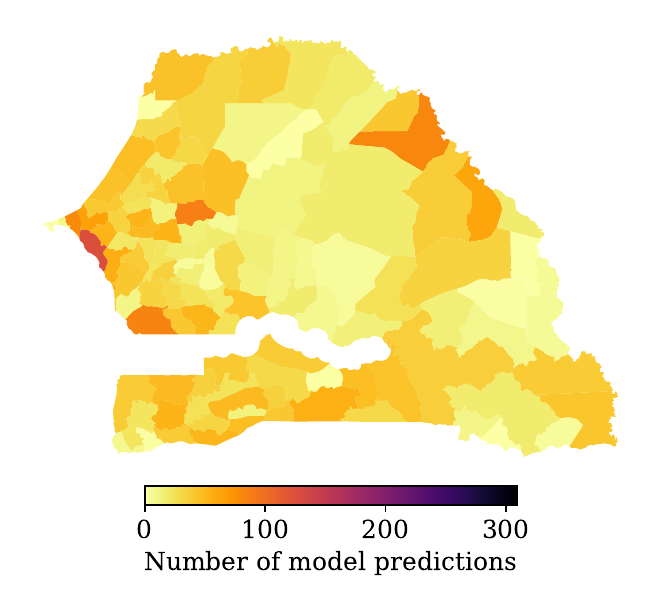}
        \caption{Additional school predictions}
    \end{subfigure}
    \caption{A comparison between the distribution of government-identified schools and model predictions across Senegal ($\tau=0.5$ and $d=250$ m). (a) Plot of the number of government-identified schools per district. (b) Plot of the number of model predictions per district. (c) Plot of the difference between the number of model predictions and government-identified schools.}
    \label{fig:choropleth}
\end{figure*}

\begin{table}[t]
    \centering
    \begin{tabular}{lcccccc}
    \toprule 
       & \textbf{F2 score}  & \textbf{Recall} & \textbf{Precision} & $\boldsymbol{\tau^*}$\\
    \midrule
       \textbf{BEN}  & 0.982 & 0.990 & 0.952 & 0.366\\
       \textbf{BWA}  & 0.960 & 0.968 & 0.929 & 0.352 \\
       \textbf{GHA}  & 0.968 & 0.985 & 0.905 & 0.386\\
       \textbf{KEN}  & 0.966 & 0.952 & 0.970 & 0.395\\
       \textbf{MWI}  & 0.953 & 0.976 & 0.872 & 0.335 \\
       \textbf{NAM}  & 0.914 & 0.922 & 0.885 & 0.315\\
       \textbf{RWA}  & 0.978 & 0.996 & 0.910 & 0.344 \\
       \textbf{SEN}  & 0.985 & 0.997 & 0.940 & 0.355 \\
       \textbf{SSD}  & 0.962 & 0.975 & 0.914 & 0.378 \\
       \textbf{ZWE}  & 0.977 & 0.992 & 0.919 & 0.327\\
    \bottomrule
    \end{tabular}
    \caption{Test set performance statistics of the best model per country at the probability threshold $\tau^*$ that maximizes the F2 score.}
    \label{tab:fscore-results}
\end{table}

\subsection{Human-in-the-loop: Model validation tool}
We developed an interactive web map using Dash \cite{plotly2024dash} and Mapbox \cite{mapbox2024} to assist government partners in discovering new, previously unmapped schools. Using the tool, users can visualize model predictions, compare them with official school datasets, and validate the predictions by either cross-referencing them against public records (e.g., OSM, Google Maps) or using the provided lat-lon points for field validation.

\subsubsection{Components}
The tool allows users to control the information shown on the map by adjusting the probability and distance thresholds, described as follows:
\begin{itemize}
    \item \textbf{Probability threshold $\boldsymbol{\tau}$}. This value determines the minimum confidence scores required for model predictions to be displayed on the map. Setting a higher probability threshold increases precision but decreases recall, resulting in fewer but more confident predictions. A lower threshold increases recall but decreases precision, leading to more predictions, many with low confidence scores and more likely to be false positives.
    \item \textbf{Distance threshold $\boldsymbol{d}$}. This value sets the maximum distance for a model prediction to ``match" a government data point (default value is 250 m). Lower values require predictions to be geographically close to the government data points to be considered a match, while higher values allow matches at greater distances. Users can set higher thresholds if government coordinates are imprecise and far from actual school buildings.
\end{itemize}
Using our tool, users can filter out model predictions that match nearby government data points (based on the distance threshold), leaving only unmatched predictions for validation. We generally recommend government partners begin with a high probability threshold and validate in decreasing order of confidence scores. To assess the effectiveness and usability of the tool, we have partnered with selected governments of our pilot countries to user-test the validation tool with frequent training and feedback sessions.

\subsubsection{Comparison with government data: A case study in Senegal.}  We illustrate in Figure \ref{fig:venn-diagram} a comparison of the model predictions (12,352) and government-registered schools (9,033) in Senegal for $\tau=0.5$ and $d=250$ m. We chose these thresholds for demonstration purposes, but in practice, users can dynamically adjust these values as needed. For the government dataset, we used the raw, unaltered GPS coordinates of the original dataset to display on the map. However, for this analysis, we disregarded all duplicate points and points in non-built-up areas in the government dataset for comparison with the model predictions.

\begin{figure}
    \centering
    \includegraphics[width=0.7\linewidth]{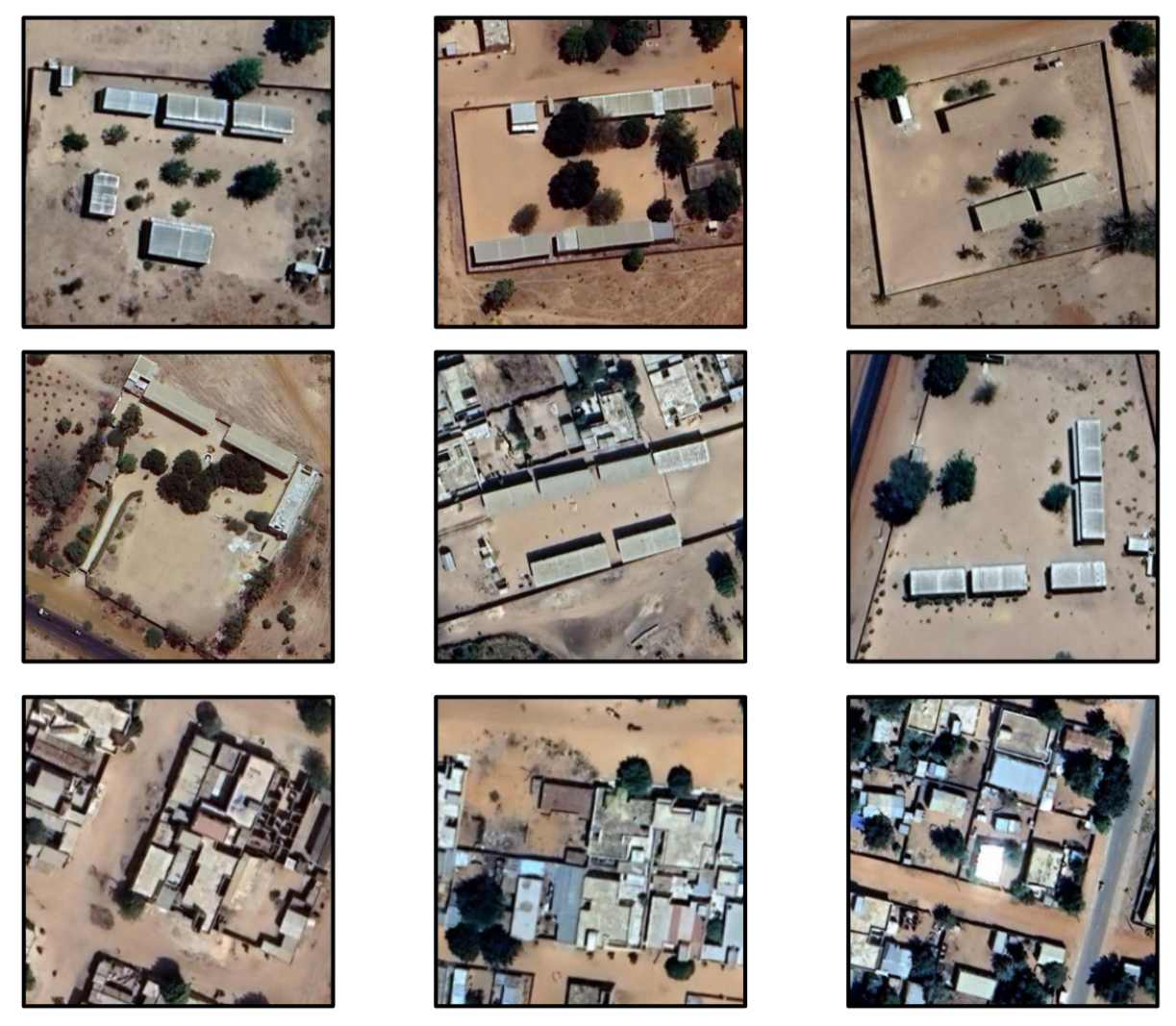}
    \caption{(Top row) Model predictions with matching government-identified schools. (Middle row) Model predictions with no matches in the government dataset. (Bottom row) Government-identified schools with no matching model prediction.}
    \label{fig:examples-results}
\end{figure}

We matched a total of 6,983 schools, leaving 2,050 unmatched schools in the government data and 5,369 unmatched school predictions to be validated. Note that the unmatched government schools may include both false negatives and erroneous points (e.g. coordinates that are far from school buildings), as shown in Figure \ref{fig:examples-results}. We also note that these figures can change by varying the probability threshold and distance threshold. This adaptable approach allows us to meet the various requirements and constraints of different government stakeholders, depending on the resources available to them for model validation.

\section{Conclusion}
We have presented an end-to-end pipeline for generating nationwide school location predictions using deep learning and high-resolution satellite images. We have shown that using only classification-level annotations, we can approximate the precise locations of schools at the level of GPS coordinates, which government partners can readily use for remote or field validation. In this work, we underscore the importance of local data collection and rigorous model evaluation to obtain the best models tailored to local contexts. We also emphasize the significance of strong government engagement for the successful adoption of AI-based mapping solutions in development contexts. Ultimately, by harnessing innovative technologies, governments and connectivity providers are better equipped to accurately estimate the cost of connecting schools and build financially sustainable solutions to fast-track nationwide school connectivity. 

In future works, we plan to analyze the government-validated model outputs and determine the extent to which these results can be used to further improve model performance \cite{monarch2021human}. We also plan to experiment with domain adaptation methods to identify schools in countries with little to no available school data \cite{song2019domain,peng2022domain}. 
The code for this work is made available at \url{https://github.com/unicef/giga-global-school-mapping}.

\section{Acknowledgments}
We gratefully acknowledge Giga, a joint initiative by UNICEF and ITU, for providing the funding and resources that made this research possible. The high-resolution satellite imagery from Maxar was generously provided through the support of the United States Government under the NextView end-user license. We also express our gratitude to Dell for granting us access to HPC clusters with NVIDIA GPU support, which were instrumental to this work.

Special thanks to Do-Hyung Kim, Naroa Zurutuza, Elena Fuestch, Lema Zekrya, Munkhkhuj Badarch, Kelsey Doerksen, and Casper Fibæk for their invaluable insights and expertise, which significantly enhanced the quality of this research.


\end{document}